%% file: main.tex
\pdfoutput=1

\documentclass[11pt]{article}

\usepackage[]{emnlp2021}

\usepackage{times}
\usepackage{latexsym}
\usepackage{header}

\usepackage[T1]{fontenc}

\usepackage[utf8]{inputenc}

\usepackage{microtype}

\title{SeaD: End-to-end Text-to-SQL Generation with Schema-aware Denoising}

\author{
    Kuan Xu
    \and Yongbo Wang
    \and Yongliang Wang
    \and Zujie Wen
    \and Yang Dong
    \\
    Ant Group \\ Hangzhou, China \\ 
    \texttt{\{xukuan.xk,wyb269207,yongliang.wyl,zujie.wzj,doris.dy\}@antgroup.com}
}

\begin{document}
\maketitle
\begin{abstract}
\input{0_abstract}
\end{abstract}

\input{1_introduction}
\input{2_related_work}
\input{3_methodology}
\input{4_experiments}
\input{5_conclusion}

\bibliography{anthology,custom,ksmo}
\bibliographystyle{acl_natbib}
\end{document}

%% file: 0_abstract.tex
In text-to-SQL task, seq-to-seq models often lead to sub-optimal performance
due to limitations in their architecture.
In this paper, we present a simple yet effective approach that adapts transformer-based seq-to-seq model to robust text-to-SQL generation.
Instead of inducing constraint to decoder or reformat the task as slot-filling, we propose to train seq-to-seq model with \underline{S}ch\underline{e}ma-\underline{a}ware \underline{D}enoising (SeaD),
which consists of two denoising objectives that train model to either recover input or predict output from two novel \emph{erosion} and \emph{shuffle} noises. 
These denoising objectives
acts as the auxiliary tasks for better modeling the structural data in S2S generation.
In addition, we improve and propose a clause-sensitive execution guided (EG) decoding strategy to overcome the limitation of EG decoding for generative model.
The experiments show that the proposed method improves the performance of seq-to-seq model in both schema linking and grammar correctness and establishes new state-of-the-art on WikiSQL benchmark\footnote{\url{https://github.com/salesforce/WikiSQL}}.
The results indicate that the capacity of vanilla seq-to-seq architecture for text-to-SQL may has been under-estimated.

%% file: 1_introduction.tex
\section{Introduction}
Text-to-SQL aims at translating natural language into valid SQL query. It enables layman to explore structural database information with semantic question instead of dealing with the complex grammar required by logical form query.
Though being a typical seq-to-seq (S2S) task, auto-aggressive models (LSTM, Transformer, etc.) that predict target sequence token by token fail to achieve state-of-the-art results for text-to-SQL.
Previous works attribute the sub-optimal results to three major limitations.
First, SQL queries with different clause order may have exact same semantic meaning and return same results by execution. The token interchangeability may confusion model that based on S2S generation.
Second, the grammar constraint induced by structural logical form is ignored during auto-aggressive decoding, therefore the model may predict SQL with invalid logical form.
Third, schema linking, which has been suggested to be the crux of text-to-SQL task, is not specially addressed by vanilla S2S model.

Recent years, there have been various works proposed to improve the performance of S2S model with sketch-based slot-filling, constraint (structural-awared) decoder or explicitly schema linking module, that try to mitigate these limitations respectively.
Though most of these works share the same encoder-decoder architecture with S2S model, they only consider it as a simple baseline.
On the other side, generative model exhibits huge potential in other tasks that also require structural output, which shares similar property with text-to-SQL.
This raised the question: \emph{Is the capacity of  generative model underestimated or not?}

\input{figures/fig_s2s}
In this paper, we investigate this question based on the Transformer architecture.
Instead of building extra module or placing constraint on model output, we propose novel schema-awared denoising objectives trained along with the original S2S task.
These denoising objectives deal with the intrinsic attribute of logical form and therefore facilitate schema linking required for text-to-SQL task.
The inductive schema-awared noises can be categorized into two types: \emph{erosion} and \emph{shuffle}.
Erosion acts on schema input by randomly permute, drop and add columns into the
current schema set.
The related schema entity in target SQL query will be jointly modified according to the erosion result.
Shuffle is applied via randomly re-ordering the mentioned entity and values in NL or SQL with respect to the schema columns.
During training procedure, shuffle is performed during monolingual self-supervision that trains model to recover original text given the noised one.
Erosion is applied to S2S task that trains model to generate corrupted SQL sequence, given NL and eroded schema as input.
These proposed denoising objectives are combined along with the origin S2S task to train a SeaD model.
In addition, to deal with the limitation of execution-guided (EG) decoding, we propose a clause-sensitive EG strategy that decide beam size with respect to the clause token that is predicted.

We compare the proposed method with other top-performing models on WikiSQL benchmark.
The results show that the performance of our model surpasses previous work and establish new state-of-the-art for WikiSQL.
It demonstrate the effectiveness of the schema-aware denoising approach and also shad lights on the importance of task-oriented denoising objective.

%% file: figures/fig_s2s.tex
\begin{figure}[!t]
\begin{center}
\includegraphics[width=0.47\textwidth]{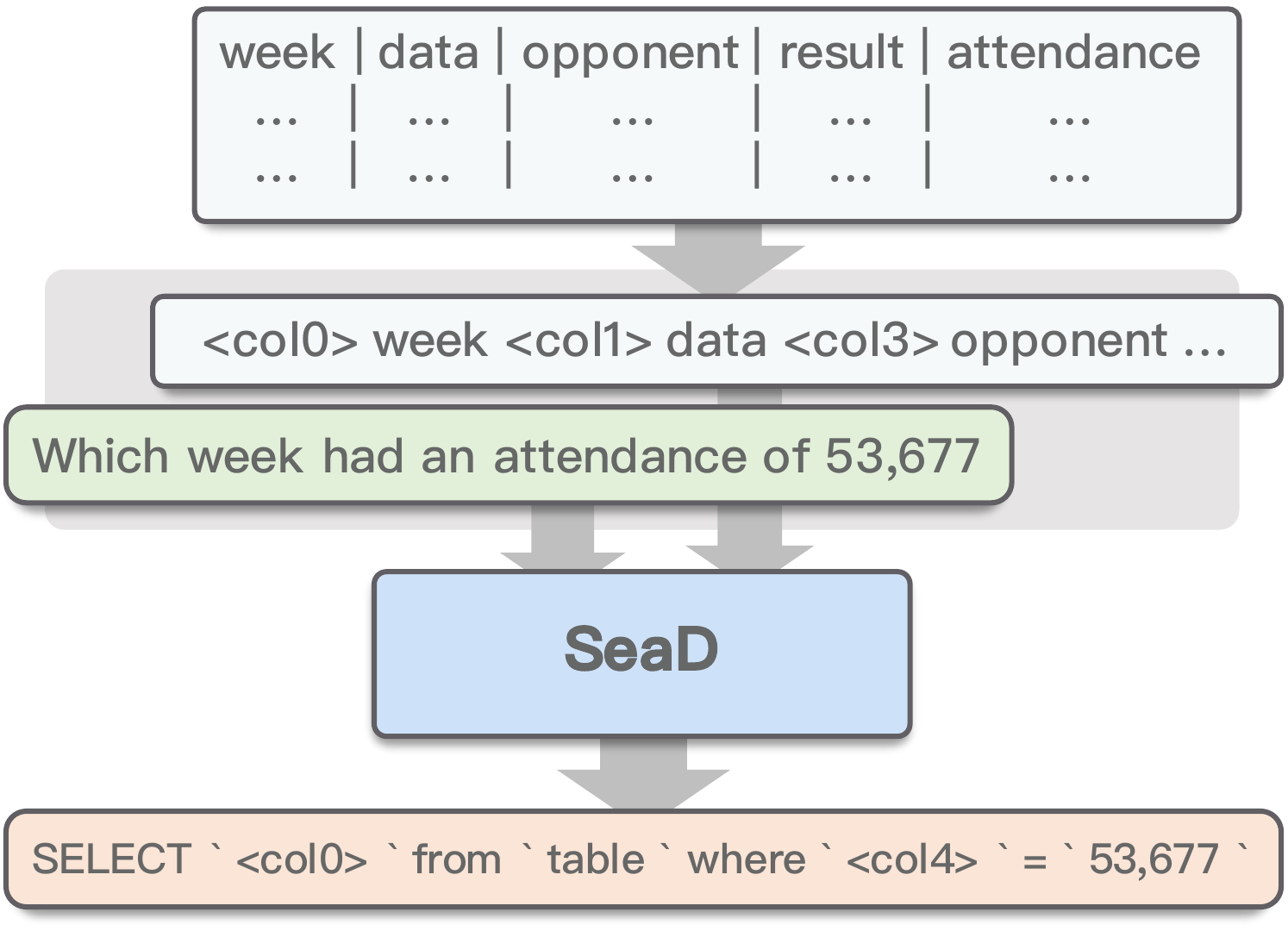}
\end{center}
\caption{\label{fig:s2s}
SeaD regards text-to-SQL as S2S generation task.
During inference, given natural language question and related database schema, SeaD directly generates corresponding SQL sequence in an auto-aggressive manner.
}
\end{figure}

%% file: 2_related_work.tex
\section{Related Work}

\textbf{Semantic Parsing}
The problem of mapping natural language to meaningful executable programs has been widely studied in natural language processing research. Logic forms \citep{zettlemoyer2012learning,artzi2011bootstrapping,artzi2013weakly,cai2013large,reddy2014large,liang2013learning,quirk2015language,chen2016latent} can be considered as a special instance to the more generic semantic parsing problem. As a sub-task of semantic parsing, the text-to-SQL problem has been studied for decades. \citep{warren1982efficient,popescu2003towards,li2006constructing,giordani2012translating,bodiksynthesizing}. Slot-filling model \citep{hwang2019comprehensive,he2019x,lyu2020hybrid} translates the clauses of SQL into subtasks, \citep{ma2020mention} treat this task as a two-stage sequence labeling model. However, the convergence rate between subtasks is inconsistent or the interaction between multiple subtasks may lead to the model may not converge well. Like lots of previous work \citep{dong2016language,lin2018nl2bash,zhong2017seq2sql,suhr2020exploring,raffel2019exploring}, we treat text-to-SQL as a translation problem, and taking both the natural language question and the DB as input.


\noindent \textbf{Hybrid Pointer Networks} Proposed by \citep{vinyals2015pointer}, copying mechanism (CM) uses attention as a pointer to copy several discrete tokens from input sequence as the output 
and have been successfully used in machine reading comprehension \citep{wang2016machine,trischler2016natural,kadlec2016text,xiong2016dynamic}, interactive conversation \citep{gu2016incorporating,yu2020online,he2019learning}, geometric problems \citep{vinyals2015pointer} and program generation \citep{zhong2017seq2sql,xu2017sqlnet,dong2016language,yu2018typesql,mccann2018natural,hwang2019comprehensive}. In text-to-SQL, CM can not only facilitate the condition value extraction from source input, 
but also help to protect the privacy of the database. In this paper, We use a Hybrid Pointer Generator Network which is similar to \citep{jia2016data,rongali2020don} to generate next step token.

\noindent \textbf{Denoising Self-training} Language model pretraining \citep{devlin2018bert,yang2019xlnet,liu2019roberta,lan2019albert} has been shown to improve the downstream performance on many NLP tasks and brought significant gains. \citep{radford2018improving,peters2018deep,song2019mass} are beneficial to S2S task, while they 
are problematic for some tasks. While \citep{lewis2019bart} is a denoising S2S pre-training model, which is effective for both generative and discriminative tasks, reduces the mismatch between pre-training and generation tasks. Inspired by this, we propose a denosing self-training architecture in training to learn mapping corrupted documents to the original.

\input{figures/denoising}

%% file: figures/denoising.tex
\begin{figure*}[!t]
\begin{center}
\begin{subfigure}[b]{0.45\textwidth}
         \centering
         \includegraphics[width=\textwidth]{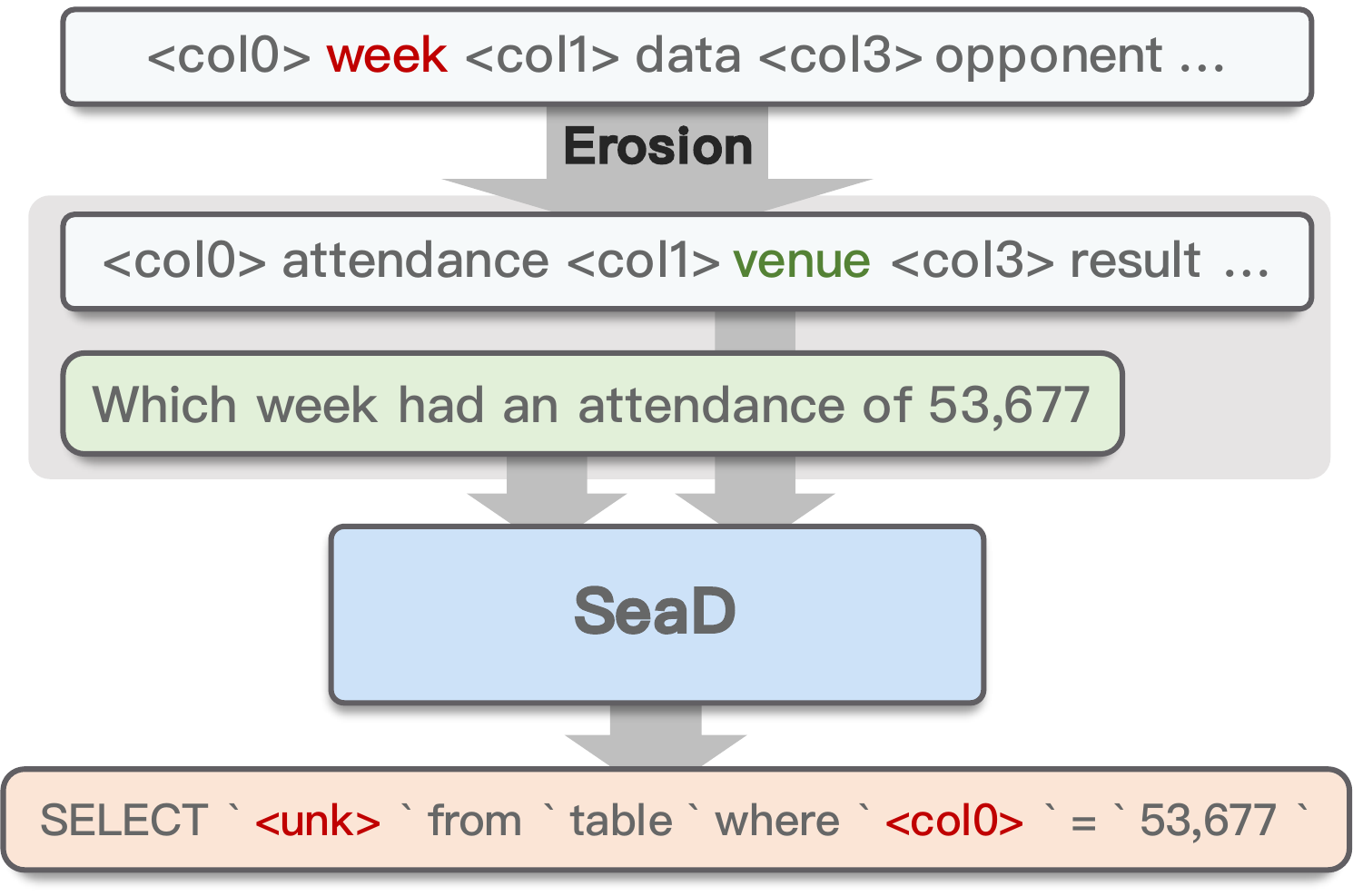}
         \caption{Erosion}
         \label{fig:erosion}
     \end{subfigure}
     \hfill
     \begin{subfigure}[b]{0.49\textwidth}
         \centering
         \includegraphics[width=\textwidth]{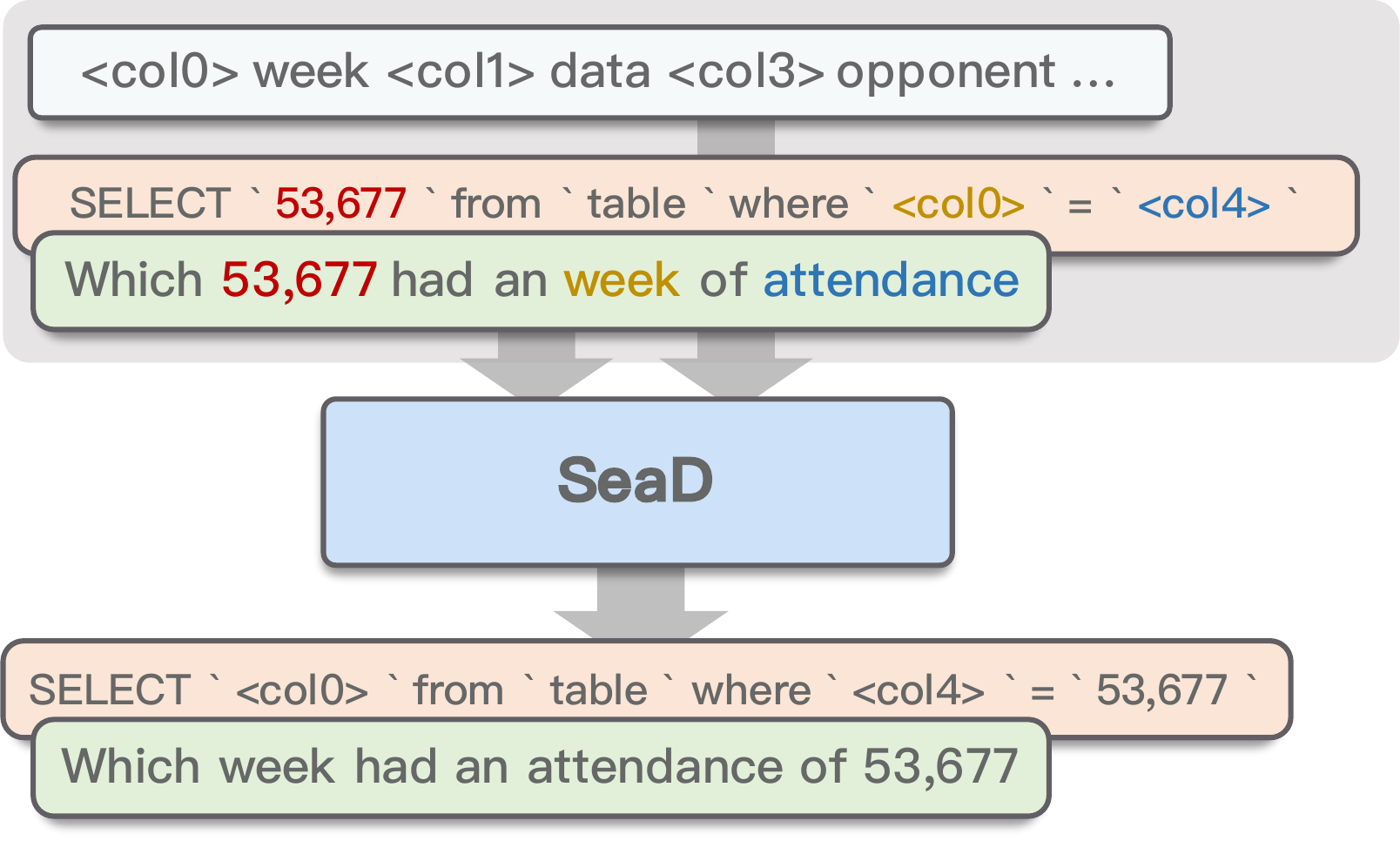}
         \caption{Shuffle}
         \label{fig:shuffle}
     \end{subfigure}
\end{center}

\caption{\label{fig:denoising}
The proposed schema-aware denoising procedure.
(a) Erosion denoising randomly drops, adds and re-permutes schema columns.
The related column entities in ground-truth SQL sequence will be jointly modified or masked out with respect to the erosion results of the current schema set. Erosion objective trains model to predict the modified SQL sequence under noised input. 
(b) Shuffle denoising objective re-permutes the mentioned entities in SQL or NL sequence, and trains model to reconstruct the sequence with the correct entity order. 
}

\end{figure*}

%% file: 3_methodology.tex
\section{Methodology}
Given natural language question $Q$ and a schema $S$, our goal is to obtain the corresponding SQL query $Y$.
Here the natural question $Q=\{q_1, ..., q_{|Q|}\}$ denotes a word sequence, the schema $S = \{\mathbf{c}_1, ..., \mathbf{c}_{|S|}\}$  is composed of a set of columns, where each column $\mathbf{c}_i=\{c_1, ..., c_{|\mathbf{c}_i|}\}$ is a sequence of words. $Y={y_1, ..., y_{|Y|}}$ denotes the token-wise raw SQL sequence.
We approach this task with directly auto-aggressive generation, i.e., predicting the SQL sequence token by token.
We choose Transformer as our base architecture, which is a widely adopted in S2S translation and generation tasks.
In this section, we first present the sample formulation that transform text-to-SQL into typical S2S task, followed by a brief introduce of the Transformer architecture with pointer generator.
Then we describe the proposed schema-aware denoising method and clause-sensitive EG decoding strategy.

\subsection{Sample Formulation}
The structural target sequence and unordered schema set require re-formulate to perform text-to-SQL task through S2S generation.
For schema formulation, each column name is prefixed with a separate special token \verb|<coli>|, where \verb|i| denotes the i-th column in the schema set.
The column type of each column is also append to the name sequence to form the  template for a schema column \verb|<coli> [col name] : [col type]|.
All columns in schema is formulated and concatenated together to obtain the representing sequence for schema.
The schema sequence is further concatenated with the NL sequence for model input.
For SQL sequence, we initialize it with raw SQL query and perform several modifications on it:
1) surrounding entities and values in SQL query with a "\verb|`|" token, and dropping other surroundings if exist;
2) replacing col entities with their corresponding separate token in schema;
3) inserting spaces between punctuation and words.
The formulated SQL sequence is illustrated in Figure~\ref{fig:s2s}.
The formatting procedure improves consistency between tokenized sequences of source and target, and contributes to the identification and linking of schema entities.

\subsection{Transformer with Pointer}
Follow the previous works on S2S semantic parsing, transformer was used in our architecture proposed by \citep{vaswani2017attention}. As we know, a complete standard SQL syntax not only needs to contain inherent keywords, but also needs to extract text span from the source input especially for condition value generation. Different from the traditional seq2seq architecture, the target words are generated from the decoder hidden states through a linear affine transformation that obtains unnormalized scores over a target vocabulary distribution. We use a Hybrid Pointer Generator Network to generate target words inspired by \citep{jia2016data}, that is say words can be picked from the target vocabulary $V=\{\mathbf{Q}_v, \mathbf{S}_v, \mathbf{V}_{sql}\}$, and words that are simply pointers to the source sequence. $\mathbf{Q}_v$ denotes words from corpus, $\mathbf{S}_v$ denotes words from column, $\mathbf{V}_{sql}$ denotes SQL keywords (like \verb|SELECT, MAX, MIN|, etc.).

In decoding process, for each input sequence $I_{source}$, we use the transformer encoder to encode it to the hidden states $H_{target}$. First, the transformer decoder produces the hidden states $h_t$ in step $t$ based on previously generated sequence and encoded output as described in \citep{vaswani2017attention}, then we use a affine transformation on $h_t$ to obtain scores $scores_{vocab}=\{s_1, ..., s_{|V|}\}$ over target vocabulary $V$ for each word. As well as we use $h_t$ to compute unnormalized attention scores $score_{source}=\{i_1, ..., i_{|input|}\}$ with the encoded sequence. Concatenating $scores_{vocab}$ and $score_{source}$ directly to get hybrid scores $score_{hybrid}=\{s_1, ..., s_{|V|}, i_1, ..., i_{|input|}\}$ like in \citep{rongali2020don}, the first $|V|$ positions are the output distribution of the target vocabulary and the last $|input|$ positions are the words pointing to the source tokens. We compute the final probability distribution by $P=\mathbf{softmax}(score_{hybrid})$. $P$ is used in loss function and next token generation respectively while training and inference.

\subsection{Schema-aware Denoising}
Similar to masked language modeling and other denoising task, we propose two schema-aware objectives, erosion and shuffle, that train model to either reconstruct the origin sequence from noising input or predict corrupted output otherwise.
The denoising procedure is illustrated in Figure~\ref{fig:denoising}.
\subsubsection{Erosion}
Given input sequence $X=\{Q, S\}$, where $Q$ denote the NL sequence, erosion corrupts the schema sequence $S$ with a serial compositions of three noising operations:

\noindent\textbf{Permutation} Re-order the concatenation sequence of schema columns during schema formulation.

\noindent\textbf{Removal} For each column, remove it with a dropping probability $p_{drop}$.

\noindent\textbf{Addition} With a addition probability $p_{add}$, extract a column from another schema that exists in the training database and insert it into current schema set.

\noindent During all operations above, the order of separating special tokens remains unchanged, therefore the corresponding anonymous entities in SQL query should be updated along with the erosion operations in schema sequence. 
In particular, if a column entity mentioned in SQL query is removed during erosion, we substitute the corresponding column token in SQL with a masking token \verb|<unk>| to cope with the absence of the schema information.
With such joint modification for schema and SQL sequence, the model is required to identify the schema entities that are truly related to the NL question and learns to raise an unknown exception whenever the schema information is insufficient to compose the target SQL.
\input{figures/algo}

\subsubsection{Shuffle}
Given input sequence $X'=\{\mathcal{Q}, S\}$, where $\mathcal{Q} = \{Q, Y\}$, the shuffle noise reorders the mentioning sequence of entities in the source query while the schema sequence $S$ is fixed.
The denoising objective trains model to reconstruct the query sequence $\mathcal{Q}$ with entities in correct order.
The objective of recovering shuffled entity orders trains model to capture the inner relation between different entities and therefore contributes to the schema linking performance.
It is also notable that, as a self-supervision objective, both $Q$ and $Y$ are engaged in this denoising task and get trained separately.
Though we dependent on the SQL query to identify the value entities in NL query, order shuffling with only column entities is sufficient to obtain promising performance.
Since no parallel data is required, additional corpus with monolingual data for both SQL and NL could help with the re-order task and will be one of the further direction of this work.

\subsubsection{Training Procedure}
Inspired by previous works on denoising self-training~\citep{song_mass_2019,lewis_bart_2019}, we propose to train the schema-aware denoising objectives along with the main S2S task.
During training, for each training sample, we apply a nosing pipeline to it before feeding it into the model.
The noises with different type are applied to the sample individually.
Through the control of activate probability, they could share the same weights in the overall objective.
Such continual noising pipeline generates random-wise corrupted samples during training.
It prevents the model from fast over-fitting and could yield results with better generalization~\citep{siddhant_leveraging_2020}.
The whole procedure is summarized in Algorithm~\ref{alg:noising}.
\subsection{Clause-sensitive EG Decoding}
During the inference of text-to-SQL task, the predicted SQL may contain errors related to inappropriate schema linking or grammar.
EG decoding~\citep{wang2018robust} is proposed to amend these errors through an executor-in-loop iteration.
It is performed by feeding SQL queries in the candidate list into the executor in sequence and discarding those queries that fail to execute or return empty result.
Such decoding strategy, while effective, suggests that the major disagreement in the candidate list focuses on schema linking or grammar.
Directly perform EG to the candidates generated with beam search leads to trivial improvement,
as the candidates consist of redundant variations focuses on selection or schema naming, etc.
This problem can be addressed by setting the beam length of most of the predicted tokens to $1$ and releasing those tokens related to schema linking (e.g., \verb|WHERE|).
We also notice that there are cases that combine incorrect schema linking with some aggregation in \verb|SELECT| clause, which return some trivial results such as $0$, thus suppress the EG filter.
To mitigate the issue, we suggest to drop aggregate operator in \verb|SELECT| during EG to maximize the effectiveness of it.
Note that with such strategy, the condition with inequation in \verb|WHERE| clause should be dropped together to ensure the validity of the ground-truth SQL results.


%% file: figures/algo.tex
\begin{algorithm}[tb]
\SetAlgoLined
    \SetKwInOut{Init}{Init}
    \SetKwInOut{Input}{Input}
    \SetKwFunction{Erosion}{Erosion}
    \SetKwFunction{Infilling}{Infilling}
    \SetKwFunction{Shuffle}{Shuffle}
    \SetKwFunction{Type}{SeqType}
    \SetKwFunction{IsSQL}{IsSQL}
    \SetKwFunction{Train}{TrainOneSample}
    \SetKwIF{With}{elsewith}{withElse}{with}{do}{else with}{else}{end}
    
    \Input{training corpus $\mathcal{X} = \{(Q_i, S_i, Y_i)\}, i\in 1, ... |\mathcal{X}|$, S2S Transformer $\Theta$}
    
    \ForEach{$(Q_i, S_i, Y_i) \in \mathcal{X}$}{
        $T_{src}, T_{tgt} \leftarrow Q_i, Y_i$\;
        $T_{tgt}, S_i \leftarrow$ {\Erosion{$T_{tgt}$, $S_i$}} \\
        \With{$P_{shuffle}$}{
            \With{$P_{swap}$}{
                $T_{src}, T_{tgt} \leftarrow T_{tgt}, T_{src}$\;
            }
            $T_{src}$ $\leftarrow$ \Shuffle($T_{tgt}$) \\
        }
        $T_{type}$ $\leftarrow$ \Type{$T_{tgt}$}\\
        \eIf{$T_{type} = \texttt{SQL}$}
        {$T_{prefix} \leftarrow \texttt{<2sql>}$\;}
        {$T_{prefix} \leftarrow \texttt{<2nl>}$\;}
        $T_{src} \leftarrow T_{prefix} + T_{src} + S_i$\;
        \Train{$T_{src}, T_{tgt}, \Theta$}
    }
\caption{\label{alg:noising}Training procedure for schema-aware denoising}
\end{algorithm}

%% file: 4_experiments.tex
\section{Experiment}
To demonstrate the effectiveness of the proposed method, we evaluate the proposed model on WikiSQL dataset and compare it to other state-of-the-art models.

\subsection{Dataset}
The WikiSQL dataset consists of $56,355$, $8,421$ and $15,878$ NL-SQL pairs for training, validation and inference respectively.
All ground-truth SQL queries are guaranteed with at least one query result.
Each SQL contains \verb|SELECT| clause with at most one aggregation operator and \verb|WHERE| clause with at most $4$ conditions that connected by \verb|AND|.
Each SQL is associated with a schema in database.

\input{figures/tb_cmp}
\subsection{Implementation details}
We implement our method using AllenNLP~\citep{gardner_allennlp_2018} and Pytorch~\citep{paszke_pytorch_2019}.
For the model architecture, we use Transformer with $12$ layers in each of the encoder and decoder with a hidden size of $1024$.
We initialize the model weight with \verb|bart-large| pretrained model provided by Huggingface community~\citep{wolf_huggingfaces_2020} and fine-tune it on training dataset for $20$ epochs.
The batch size during training is set to $8$ with a gradient accumulation step of $2$.
We choose Adam~\citep{kingma_adam:_2014} as the optimizer and set the learning rate to $7e-5$ with a  warm-up step ratio of $1\%$.
The weight decay for regulation is set to $0.01$.
We set the activation probability $P_{swap} = 0.5$ and $P_{shuffle} = 0.3$ to balance the weight between self-supervision and S2S objective.
$P_{drop}$ for column removal in erosion is set to $0.1$.
The early stop patience is set to $5$ with respect to the BLUE metric~\citep{papineni_bleu_2001} on validation set.
The overall training procedure spend around $3$ hours on an Ubuntu server with $8$ NVIDIA V100 GPUs.

\subsection{Competitors}
We compare the proposed method to the following models: (1) SQLnet \citep{xu2017sqlnet} is a sketch-based method; (2) SQLova \citep{hwang2019comprehensive} is a sketch-based method which integrates the pre-trained language model; (3) X-SQL \citep{he2019x} enhances the structural schema representation with the contextual embedding; (4) HydraNet \citep{lyu2020hybrid} uses column-wise ranking and decoding; (5) IESQL \citep{ma2020mention} treats Text-to-SQL as sequence labeling based model; (6) BRIDGE \citep{lin2020bridging} is a sequential architecture for modeling dependencies between natural language question and relational DBs; (7) SDSQL \citep{hui2021improving} is a multi-task model with schema dependency guided.

\subsection{Comparison with State-of-the-art Models}
The comparison results are summarized in Table~\ref{tb:cmp}.
Models suffixed with $\clubsuit$ leverage additional annotation of the dataset.
Models suffixed with $\diamondsuit$ utilize database content during training procedure.
Without using EG, SeaD significantly outperforms all models without the auxiliary of table content or schema linking annotation.
When combined with EG decoding, SeaD achieve best performance even compared to those models that utilize additional training information.
It indicates the effectiveness of the proposed denoising objectives on modeling text-to-SQL through vanilla S2S.
\input{figures/tb_cmp_eg}
Notably, the annotation noise makes aggregation prediction a major challenge for WikiSQL.
Previous works suggested to improve AGG prediction via rule-based annotation amendment.
As shown in Table~\ref{tb:cmp_eg}, we argue that the proposed aggregation dropping strategy for EG achieves comparable enhancement, while less human effort is involved.
Combined with the AGG dropped clause-sensitive EG, the SeaD model establishes new state-of-the-art on WikiSQL benchmark.

To analysis the detailed improvement for SeaD on text-to-SQL task, in Table~\ref{tb:clause-err} we report the accuracy on WikiSQL test set with respect to several SQL components with and without EG decoding.
SeaD shows promising results on column selection, aggregation, where column and where value prediction.
It outperforms all method except SDSQL, which leverages rule-based annotation of schema linking.
After applying EG decoding, SeaD achieves best performance on four out of five components among all competitors.
\input{figures/tb_clause_err}
\subsection{Ablation Study}
\input{figures/tb_ablation}
To evaluate the contribution of each proposed objective, we perform ablation study to SeaD (\ref{tb:ablation}) with WikiSQL dataset.
We start from the Bart model and add components to it in sequence.
The pointer net contributes to $1.2\%$ absolute improvement of $Acc_{lf}$ on test set.
Combine text infilling, an effective denoising objective utilized by Bart, into training procedure brings $0.3$ absolute $Acc_{lf}$ improvement.
On the other hand, erosion and shuffle objectives contribute to $1.5\%$ and $0.6\%$ absolute $Acc_{lf}$ improvement for SeaD on test set respectively.
It demonstrates the effectiveness of the schema-aware denoising objective for improving S2S generation in text-to-SQL task.

%% file: figures/tb_cmp.tex
\begin{table}
\centering
\scalebox{0.9}{
\begin{tabular}{lcccc} %
\toprule

\textbf{Model} & \multicolumn{2}{c}{\textbf{Dev}} &
\multicolumn{2}{c}{\textbf{Test}} \\ 
\cmidrule{2-3} \cmidrule{4-5}

 & $Acc_{lf}$ & $Acc_{ex}$ & $Acc_{lf}$ & $Acc_{ex}$\\
\midrule
SQLNet              & 63.2 & 69.8 & 61.3 & 68.0 \\
SQLova              & 81.6 & 87.2 & 80.7 & 86.2 \\
X-SQL               & 83.8 & 89.5 & 83.3 & 88.7 \\
HydraNet            & 83.6 & 89.1 & 83.8 & 89.2 \\
SeaD                & 84.9 & 90.2 & 84.7 & 90.1 \\
IESQL~$^\clubsuit$     & 84.6 & 89.7 & 84.6 & 88.8 \\
BRIDGE~$^\diamondsuit$ & 86.2 & 91.7 & 85.7 & 91.1 \\
SDSQL~$^\clubsuit$     & \textbf{86.0} & \textbf{91.8} & \textbf{85.6} & \textbf{91.4} \\
\midrule
HydraNet+EG             & 86.6 & 92.4 & 86.5 & 92.2 \\
IESQL+EG~$^\clubsuit$      & 85.8 & 91.6 & 85.6 & 91.2 \\
BRIDGE+EG~$^\diamondsuit$  & 86.8 & 92.6 & 86.3 & 91.9 \\
SDSQL+EG~$^\clubsuit$      & 86.7 & 92.5 & 86.6 & 92.4 \\
SeaD+EG$_{CS}$     & \textbf{87.3} & \textbf{92.8} & \textbf{87.1} & \textbf{92.7} \\
\bottomrule
\end{tabular}
}
\caption{\label{tb:cmp}
Accuracy ($\%$) of logic form ($Acc_{lf}$) and execution ($Acc_{ex}$) of our model SeaD and other competitors.
Best results in bold.
EG: execution-guided decoding.
EG$_{CS}$: the proposed clause-sensitive EG strategy for S2S generation.
$\clubsuit$ denotes methods that leverage additional annotation of dataset.
$\diamondsuit$ denotes methods that utilize database content during training.
}
\end{table}

%% file: figures/tb_cmp_eg.tex
\begin{table}
\centering
\scalebox{0.9}{
\begin{tabular}{lcccc} %
\toprule
\textbf{Model} & \multicolumn{2}{c}{\textbf{Dev}} &
\multicolumn{2}{c}{\textbf{Test}} \\ 
\cmidrule{2-3} \cmidrule{4-5}

 & $Acc_{lf}$ & $Acc_{ex}$ & $Acc_{lf}$ & $Acc_{ex}$\\
\midrule
IESQL+EG+AE & \textbf{87.9} & 92.6 & \textbf{87.8} & 92.5 \\
SDSQL+EG+AE & 86.7 & 92.5 & 87.0 & 92.7 \\
SeaD+EG$_{ACS}$ & 87.6 & \textbf{92.9} & {87.5} & \textbf{93.0} \\
\bottomrule
\end{tabular}
}
\caption{\label{tb:cmp_eg}
Accuracy ($\%$) of logic form ($Acc_{lf}$) and execution ($Acc_{ex}$) of our model SeaD and other competitors with EG decoding.
Best results in bold.
EG: execution-guided decoding.
AE: rule-based aggregation enhancement.
EG$_{ACS}$: the clause-sensitive EG strategy for S2S generation, with aggregation ignored during decoding.
}
\end{table}

%% file: figures/tb_clause_err.tex
\begin{table}
\centering
\scalebox{0.92}{
\begin{tabular}{lccccc} %
\toprule
Model & $S_{col}$ & $S_{agg}$ & $W_{col}$ & $W_{op}$ & $W_{val}$\\
\midrule
SQLova          & 96.8 & 90.6 & 94.3 & 97.3 & 95.4\\
X-SQL           & 97.2 & 91.1 & 95.4 & 97.6 & 96.6\\
HydraNet        & 97.6 & 91.4 & 95.3 & 97.4 & 96.1\\
IESQL           & 97.6 & 90.7 & 96.4 & \textbf{98.7} & 96.8\\
SeaD            & \textbf{97.7} & \textbf{91.7} & 96.5 & 97.7 & 96.7\\
SDSQL           & 97.3 & 90.9 & \textbf{98.1} & 97.7 & \textbf{98.3}\\
\midrule
SQLova+EG       & 96.5 & 90.4 & 95.5 & 95.8 & 95.9\\
X-SQL+EG        & 97.2 & 91.1 & 97.2 & 97.5 & 97.9\\
HydraNet+EG     & 97.6 & 91.4 & 97.2 & 97.5 & 97.6\\
IESQL+EG        & 97.6 & 90.7 & 97.9 & \textbf{98.5} & 98.3\\
SeaD+EG$_{CS}$  & \textbf{97.9} & \textbf{91.8} & \textbf{98.3} & 97.9 & \textbf{98.4}\\
\bottomrule
\end{tabular}
}
\caption{\label{tb:clause-err}
Test accuracy ($\%$) on WikiSQL test set for various clause components of SQL.
The best results in bold.
EG: execution-guided decoding.
EG$_{CS}$: clause-sensitive EG decoding for S2S generation.
}
\end{table}

%% file: figures/tb_ablation.tex
\begin{table}
\centering
\scalebox{.85}{
\begin{tabular}{lcccc} %
\toprule
\textbf{Model} & \multicolumn{2}{c}{\textbf{Dev}} &
\multicolumn{2}{c}{\textbf{Test}} \\ 
\cmidrule{2-3} \cmidrule{4-5}

 & $Acc_{lf}$ & $Acc_{ex}$ & $Acc_{lf}$ & $Acc_{ex}$\\
\midrule
Bart                            & 81.4 & 87.1 & 81.2 & 86.8 \\
Bart$_{ptr}$                    & 82.8 & 88.6 & 82.4 & 88.3 \\
Bart$_{ptr}$ + infilling        & 82.8 & 88.7 & 82.7 & 88.6 \\
SeaD (Shuffle-only)             & 83.5 & 89.0 & 83.2 & 88.8 \\
SeaD (Erosion-only)             & 84.2 & 89.6 & 84.1 & 89.4 \\
SeaD                            & 84.6 & 90.2 & 84.7 & 90.1 \\
\bottomrule
\end{tabular}
}
\caption{\label{tb:ablation}
Ablation study for SeaD model on WikiSQL benchmark.
}
\end{table}

%% file: 5_conclusion.tex
\section{Conclusions}
In this paper, we proposed to train model with novel schema-aware denoising objectives, which could improve performance of S2S generation for text-to-SQL task.
The proposed SeaD model outperforms previous works and achieves state-of-the-art performance on WikiSQL benchmark.
The success of the SeaD highlights the potential of utilizing task-oriented denoising objective for S2S model enhancement.